\documentclass[runningheads]{llncs}
\usepackage[T1]{fontenc}
\usepackage{graphicx,verbatim}
\usepackage{marvosym}
\usepackage{cite}
\usepackage{amsmath}
\usepackage{amssymb}
\usepackage{color}
\usepackage{url}
\usepackage{hyperref}

\usepackage{booktabs}
\usepackage{multirow}
\usepackage{siunitx}
\usepackage{tabularx}
\begin{document}
\title{Speech-Guided Multimodal Learning for Vocal Tract Segmentation in Real-Time MRI}  
\titlerunning{Multimodal Learning for Vocal Tract Segmentation}


%
\author{Daiqi Liu\inst{1}(\Letter) \and
Lukas Mulzer\inst{1} \and
Md Hasan\inst{1} \and
Nyvenn de Castro\inst{2} \and
Fangxu Xing\inst{3} \and
Xingjian Kang\inst{4} \and
Chengze Ye\inst{1} \and
Siyuan Mei\inst{1} \and
Yipeng Sun\inst{1} \and\\
Tom\'{a}s Arias-Vergara\inst{1,6} \and
Jana Hutter\inst{5} \and
Jonghye Woo\inst{3} \and
Andreas Maier\inst{1} \and
Paula Andrea P\'{e}rez-Toro\inst{1,6}
}
\authorrunning{D. Liu et al.}
\institute{Pattern Recognition Lab, Friedrich\mbox{-}Alexander\mbox{-}Universit\"at Erlangen\mbox{-}N\"urnberg, Erlangen, Germany\\
\email{daiqi.deutschfau.liu@fau.de}
\and
Smart Imaging Lab, Friedrich\mbox{-}Alexander\mbox{-}Universit\"at Erlangen\mbox{-}N\"urnberg, \\
Erlangen, Germany
\and
Harvard Medical School / Massachusetts General Hospital, Boston, USA
\and
Center for AI and Data Science, Julius\mbox{-}Maximilians\mbox{-}Universit\"at W\"urzburg, W\"urzburg, Germany
\and
Institute for Information Processing, Leibniz University Hannover,\\
Hannover, Germany
\and
GITA Lab, Facultad de Ingeniería. Universidad de Antioquia UdeA,\\
Medellín, Colombia\\
}


  
\maketitle              
\begin{abstract}
Segmenting vocal tract articulators in real-time MRI (rtMRI) is a challenging dynamic image segmentation problem characterized by low contrast, rapid motion, and limited spatial resolution. However, while rtMRI acquisitions may provide synchronized acoustic signals, existing methods discard this information, and the few multimodal approaches that incorporate audio cannot be deployed when audio is unavailable. We propose a three-stage framework that leverages acoustic and phonological supervision during training while requiring only the rtMRI image at inference: phonological representations are converted into spatial bounding-box priors for articulator localization, visual and acoustic encoders are aligned via dual-level cross-modal contrastive pretraining, and the learned representations are fused through a cross-attention decoder, effectively transferring multimodal knowledge into a single-modality inference pipeline. Evaluated on 75-Speaker~Annot-16 and USC-TIMIT datasets, our method outperforms existing unimodal and multimodal methods, demonstrating that multimodal supervision provides transferable benefits for precise and clinically deployable vocal tract segmentation.


\end{abstract}
\section{Introduction}
Real-time MRI (rtMRI)  enables simultaneous, non-invasive visualization of all vocal tract articulators during continuous natural speech, making it an indispensable tool for both phonetic research and clinical speech analysis~\cite{toutios2016advances}. Accurate segmentation of these structures is critical for a wide range of clinical and translational applications, including pre-surgical planning for glossectomy, assessment of structural speech disorders (e.g., cleft lip and palate), monitoring articulatory decline in neurodegenerative diseases such as ALS, and constructing articulatory representations for speech neuroprosthetics and rehabilitation~\cite{lammert2016investigation,hagedorn2014characterizing,browman1992articulatory}.

Significant progress in medical image segmentation has been driven by deep learning, from the seminal U-Net~\cite{ronneberger2015u} and its variants to Transformer-based architectures such as Swin-UNETR~\cite{hatamizadeh2021swin} and vision foundation models including SAM~\cite{kirillov2023segment} and its medical adaptations~\cite{ma2024segment}. State space models such as U-Mamba have further extended sequential modeling capabilities to volumetric and high-resolution medical images~\cite{ma2024u}. Within the specific domain of vocal tract segmentation, early efforts to extract articulatory contours largely relied on manual or semi-automated boundary tracing, where air–tissue boundaries are hand-annotated in a reference frame, followed by nonlinear optimization propagating contours across subsequent frames~\cite{ramanarayanan2018analysis}. Other approaches assigned labels by examining pixel intensities along predefined gridlines overlaid on MR images, or by employing active shape models to capture the expected anatomical contours~\cite{labrunie2018automatic}. This process is not only time-consuming and labor-intensive but also susceptible to error, requiring extensive human supervision. Subsequent deep learning approaches applied fully convolutional networks to air-tissue boundary labeling~\cite{somandepalli2017semantic, mannem2019air} and U-Net-based architectures to multi-structure segmentation, achieving a Dice score of 0.85~\cite{ruthven2021deep}. Most recently, a multimodal framework combining rtMRI with synchronized speech audio via Transformer-based fusion established a new benchmark in speaker-independent vocal tract segmentation~\cite{jain2024multimodal}.

Despite these advances, existing multimodal approaches rely on implicit feature concatenation without explicitly modeling the correspondence between visual and acoustic representations, overlooking the complementary role of phonological class features. More critically, these methods require audio at inference time, yet clinical deployment frequently faces modality absence: conventional metallic microphones cannot be used inside MRI scanners, while fiber-optic alternatives remain prohibitively expensive; furthermore, target patient populations such as those with tongue cancer or dysarthria often present with severe speech disorders, making clean audio acquisition unreliable or impossible.

To address these limitations, we propose a three-stage multimodal framework that integrates visual, acoustic, and phonological information for precise articulator segmentation in rtMRI. Stage~1 converts phonological class descriptors into subject-specific spatial bounding-box priors, providing explicit localization guidance for small and low-contrast structures. Stage~2 pretrains the visual and audio encoders via a dual-level cross-modal contrastive objective, establishing a shared semantic space through both global and local alignment. Stage~3 fuses the aligned representations through a cross-attention decoder, in which visual tokens dynamically attend to temporally resolved audio features for fine-grained multimodal integration. Critically, the learned cross-modal representations are fully encoded into the model weights at training time, such that only the rtMRI image is required at inference, which makes our method directly deployable in clinical settings where audio may be absent or degraded. Experiments on the 75-Speaker Annot-16~\cite{shi202575} and USC-TIMIT~\cite{narayanan2014real} datasets demonstrate state-of-the-art performance, confirming that structured phonological supervision and explicit cross-modal alignment provide transferable benefits that persist at inference even without audio input.

\section{Methods}
\begin{figure}[t]
\centering
\includegraphics[width=\linewidth]{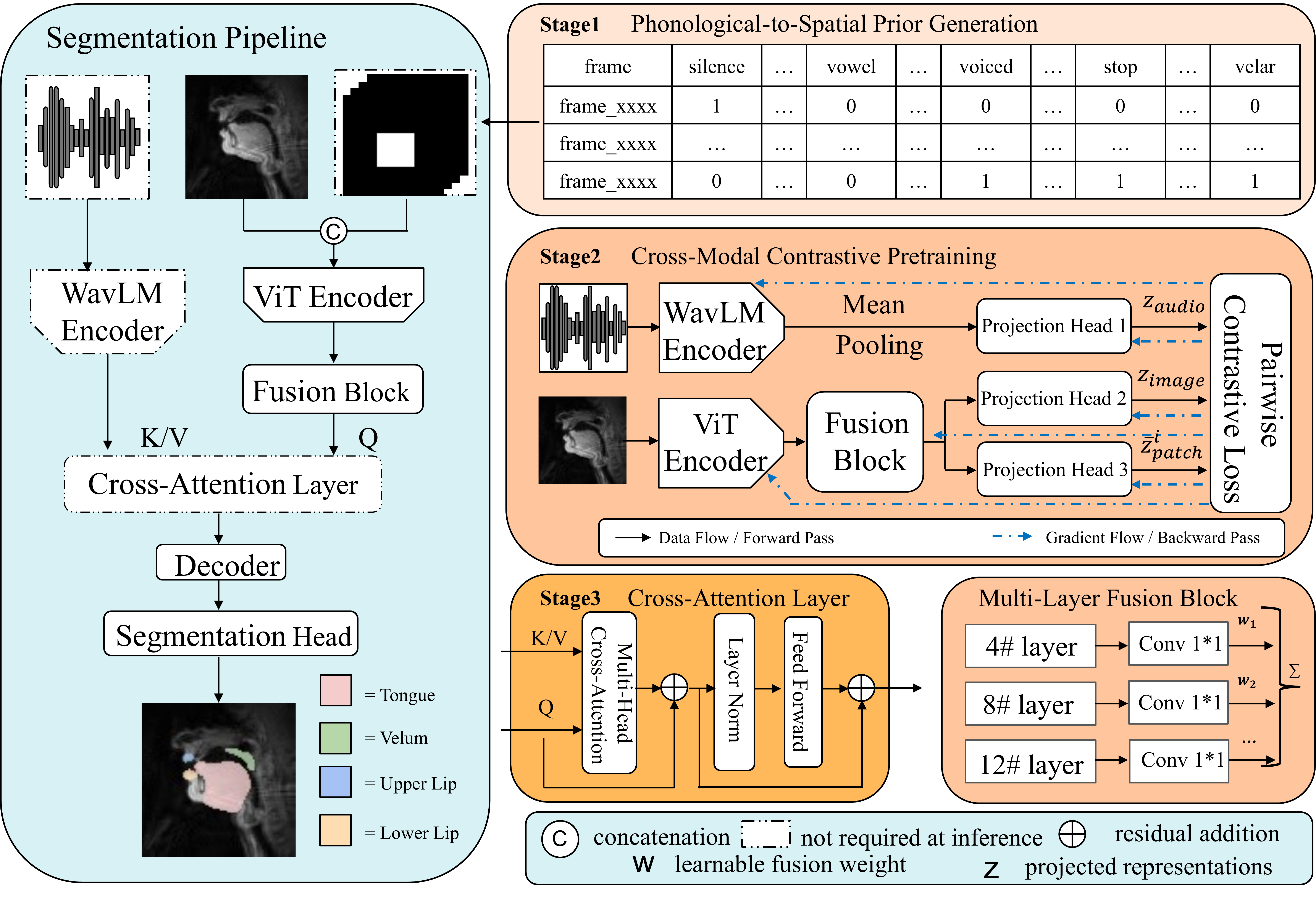} 
\caption{Schematic overview of the proposed multimodal segmentation framework. Left: The segmentation pipeline operates with three input modalities during training (rtMRI image, audio, and phonological bounding-box prior) but requires only the image at inference time. Right: Three training stages are illustrated from top to bottom. }
\label{fig:framework}
\end{figure}

As illustrated in Fig.~\ref{fig:framework}, Our method is structured into three stages that progressively incorporate phonological, acoustic, and visual information to achieve precise articulator delineation in rtMRI. Crucially, only the image is required at inference.

\subsection{Stage~1: Phonological-to-Spatial Prior Generation}
Phonological descriptors encode structured linguistic knowledge about articulatory configurations across three dimensions: \textbf{voicing} (whether vocal folds vibrate during sound production), \textbf{manner of articulation} (how airflow is shaped in the vocal tract), and \textbf{place of articulation} (where in the vocal tract the constriction occurs), providing symbolic categorical representations that explicitly describe how speech sounds are produced. For instance, the phoneme /p/ can be characterized as [voiceless, stop, labial], directly indicating the involvement of lip closure. Rather than incorporating these descriptors directly as text-based semantic features, given a multi-hot encoded phonological descriptor $\mathbf{P} \in \{0,1\}^{15}$, we generate a four-channel spatial prior by mapping each phonological class to the bounding box of its associated articulatory region. Formally, for each articulator channel $c$, the prior is computed as:
\begin{equation}
    \mathbf{x}_{\text{bbox}}^{(c)} = \text{BBox}\!\left(\textstyle\bigcup_{i:\, p_i^{(c)}=1} \mathbf{S}_i^{(c)}\right) \in \mathbb{R}^{H \times W}
\end{equation}
where $\mathbf{S}_i^{(c)}$ denotes the ground-truth segmentation mask of articulator $c$ in frame $i$, and the union is taken over all frames sharing the same phonological attribute. The operator $\text{BBox}(\cdot)$ encloses the union region with its minimum bounding rectangle, yielding an interpretable spatial prior for each of the four articulators (tongue, velum, upper lip, lower lip). To accommodate inter-speaker anatomical variability, priors are computed on a subject-specific basis.
\subsection{Stage~2: Cross-Modal Contrastive Pretraining}
Given a synchronized pair $\{\mathbf{I}, \mathbf{A}\}$, where $\mathbf{I} \in \mathbb{R}^{1 \times H \times W}$ is an rtMRI frame and $\mathbf{A} \in \mathbb{R}^{T_a}$ is the aligned audio segment, we extract modality-specific features using two pretrained encoders: 
ViT-Base~\cite{wu2020visual, deng2009imagenet} ($E_I$) for visual input and 
WavLM-Base~\cite{chen2022wavlm} ($E_A$) for audio, both with hidden dimension $D{=}768$. From $E_I$, patch tokens (excluding the \texttt{[CLS]} token) from layers $\{4, 8, 12\}$ 
are fused via learnable $1{\times}1$ convolutions to obtain multi-scale local features.
\begin{equation}
    \mathbf{Z}_{\text{patch}} = \mathcal{R}^{-1}\!\left(
    \textstyle\sum_{l \in \{4,8,12\}} 
    \mathbf{W}_l * \mathcal{R}\!\left(E_I^{(l)}(\mathbf{I})[1{:}]\right)
    \right) \in \mathbb{R}^{N \times D}
\end{equation}
where $\mathcal{R}(\cdot): \mathbb{R}^{N \times D} \to \mathbb{R}^{D \times \frac{H}{P} \times \frac{W}{P}}$ 
denotes the reshape operation converting patch tokens to spatial feature maps, 
$\mathcal{R}^{-1}(\cdot)$ is its inverse. The final-layer \texttt{[CLS]} token serves as the global visual representation  $\mathbf{Z}_{\text{image}} \in \mathbb{R}^{D}$. Concurrently, the audio feature sequence is mean-pooled along the temporal dimension to obtain a compact global representation. All representations are projected into a shared 256-dimensional space via two-layer MLPs. We enforce dual-level alignment using 
the InfoNCE loss~\cite{oord2018representation}. Global-to-global alignment matches \texttt{[CLS]} tokens with audio ($\mathcal{L}_{\text{g2g}} = \mathcal{L}(\mathbf{z}_{\text{image}}, \mathbf{z}_{\text{audio}})$), while local-to-global alignment 
enforces consistency between mean-pooled patch features and audio ($\mathcal{L}_{\text{l2g}} = \mathcal{L}(\bar{\mathbf{z}}_{\text{patch}}, \mathbf{z}_{\text{audio}})$). The total contrastive loss is defined as 
$\mathcal{L}_{\text{contrast}} = \mathcal{L}_{\text{g2g}} + 
\lambda \mathcal{L}_{\text{l2g}}$, where $\lambda = 0.5$.



\subsection{Stage~3: Cross-Attention-Based Multimodal Segmentation}
The rtMRI frame $\mathbf{I}$ and the phonological bounding-box map $\mathbf{x}_{\text{bbox}}$ from Stage~1 are concatenated channel-wise and processed through the Stage~2 pretrained ViT encoder and fusion block to obtain visual patch tokens $\mathbf{F}_v \in \mathbb{R}^{N \times D}$. Concurrently, WavLM encodes the audio into a full temporal sequence $\mathbf{F}_a \in \mathbb{R}^{T \times D}$ without pooling. A six-layer cross-attention decoder refines visual tokens by attending to audio features at each layer:
\begin{equation}
    \mathbf{F}_v^{(\ell)} = \text{LayerNorm}\left(\mathbf{F}_v^{(\ell-1)} + 
    \text{MultiHeadAttn}\left(\mathbf{F}_v^{(\ell-1)},\, \mathbf{F}_a,\, 
    \mathbf{F}_a\right)\right),
\end{equation}
where visual tokens serve as queries and audio tokens as keys and values. The refined tokens are upsampled through two convolutional blocks ($768{\to}256{\to}128$ channels) with bilinear interpolation ($14{\times}14{\to}224{\times}224$), followed by a $1{\times}1$ convolution yielding class logits $\mathbf{Y} \in \mathbb{R}^{4 \times H \times W}$. 
Training minimizes the combined loss $\mathcal{L}_{\text{seg}} = \mathcal{L}_{\text{CE}}(\mathbf{Y}, 
\mathbf{S}_{\text{gt}}) + \mathcal{L}_{\text{Dice}}(\mathbf{Y}, \mathbf{S}_{\text{gt}})$.
\section{Experiments}
\subsection{Dataset \& Preprocessing}

\noindent\textbf{75-Speaker Annot-16}
This dataset provides synchronized rtMRI videos and audio recordings from 16 speakers (7F, 9M; 8 native and 8 non-native American English speakers), with phonetic alignments and handmade articulator contour annotations. The rtMRI was acquired at 83.28 fps with  2.4 mm in-plane resolution ($84{\times}84$ pixels) using a 1.5 T scanner; audio was recorded at 20 kHz via a fiber-optic microphone. We use 12 speakers for training (51,970 frames) and 5 held-out speakers for evaluation, organized into three configurations: Seen Speaker, Unseen Task (SS-UT), Unseen Speaker, Seen Task (US-ST), and Unseen Speaker, Unseen Task (US-UT) (9,148 frames). 

\noindent\textbf{USC-TIMIT}
This dataset comprises recordings from 10 native American English speakers (5M, 5F), each producing 460 phonetically balanced sentences, acquired at 23.18 fps with 2.9 mm isotropic resolution ($68{\times}68$ pixels) using a 1.5 T scanner. Synchronized speech was simultaneously recorded using a fiber-optic microphone. Articulator contours were extracted using the semi-automatic segmentation tool provided with the dataset. We select two held-out speakers (1M, 1F) for evaluation under the US-UT configuration (5,731 frames). 

\noindent\textbf{Preprocessing}
We apply the same pipeline to both datasets. The rtMRI videos were temporally downsampled to 15 fps, from which individual frames were extracted, resized to $224 \times 224$ pixels via bilinear interpolation, and normalized per subject to $[0, 1]$ based on subject-specific minimum and maximum intensity values. Synchronized audio was resampled to 16 kHz. For each extracted frame, a three-window audio context was constructed by concatenating the aligned central segment with its preceding and following windows. Phoneme labels are mapped to 15-dimensional multi-hot phonological feature vectors~\cite{liu2025audio, arias2024contrastive}, and articulator contours were converted to pixel-level binary masks and reviewed by a speech expert, yielding four articulator classes: tongue, velum, upper lip, and lower lip.

\begin{table*}[!htpb]
\centering
\caption{Performance comparison on 75-Speaker Annot-16 and USC-TIMIT datasets. The best results are highlighted in \textbf{bold}, and the second-best are \underline{underlined}.}
\label{tab:performance_comparison}
\setlength{\tabcolsep}{3pt}
\fontsize{8}{9}\selectfont
\resizebox{\textwidth}{!}{%
\begin{tabular}{l cccccccc}
\toprule
\addlinespace[2pt]
\multirow{3}{*}{\textbf{Methods}} &
\multicolumn{6}{c}{\textbf{75-Speaker Annot-16}} &
\multicolumn{2}{c}{\textbf{USC-TIMIT}} \\
\noalign{\vskip-2.0pt}
\cmidrule(lr){2-7}\cmidrule(lr){8-9}
&
\multicolumn{2}{c}{SS-UT} &
\multicolumn{2}{c}{US-ST} &
\multicolumn{2}{c}{US-UT} &
\multicolumn{2}{c}{US-UT} \\
\noalign{\vskip-2.0pt}
\cmidrule(lr){2-3}\cmidrule(lr){4-5}\cmidrule(lr){6-7}\cmidrule(lr){8-9}
&
DSC(\%)$\uparrow$ & ASD(mm)$\downarrow$ &
DSC(\%)$\uparrow$ & ASD(mm)$\downarrow$ &
DSC(\%)$\uparrow$ & ASD(mm)$\downarrow$ &
DSC(\%)$\uparrow$ & ASD(mm)$\downarrow$ \\
\midrule

\multicolumn{9}{c}{\textbf{State-of-The-Art Unimodal Methods}} \\
\midrule
\addlinespace[1pt]

nnU-Net~\cite{isensee2021nnu} &
$74.59\pm10.09$ & $5.66\pm4.79$ & $73.11\pm11.56$ & $5.90\pm7.67$ & $69.77\pm10.49$ & $6.87\pm8.58$ & $66.55\pm13.27$ & $7.81\pm9.35$ \\
\addlinespace[2pt]

ResUNet-a~\cite{diakogiannis2020resunet} &
$69.24\pm12.14$ & $6.13\pm5.27$ & $68.27\pm10.43$ & $6.60\pm5.77$ & $67.58\pm11.27$ & $7.41\pm9.35$ & $65.67\pm12.78$ & $7.66\pm8.29$ \\
\addlinespace[2pt]

Swin-UNETR~\cite{hatamizadeh2021swin} &
$72.38\pm10.58$ & $5.22\pm4.91$ & $72.72\pm9.11$ & $5.06\pm6.98$ & $70.61\pm10.93$ & $4.37\pm6.20$ & $69.40\pm11.19$ & $4.86\pm7.54$ \\
\addlinespace[2pt]

MedSAM-2~\cite{zhu2024medical} &
$75.14\pm8.93$ & $3.11\pm3.96$ & $73.29\pm8.57$ & $3.17\pm4.10$ & $71.57\pm9.12$ & $3.50\pm4.51$ & $69.22\pm9.61$ & $4.48\pm5.46$ \\
\addlinespace[2pt]

MedSegDiff-V2~\cite{wu2024medsegdiff} &
$69.01\pm12.47$ & $5.10\pm3.44$ & $67.04\pm18.18$ & $5.08\pm3.14$ & $65.43\pm13.09$ & $6.71\pm5.17$ & $63.64\pm13.28$ & $5.59\pm4.20$ \\
\addlinespace[2pt]

DINOv2~\cite{oquab2023dinov2} &
$77.83\pm11.53$ & $3.76\pm4.55$ & $77.70\pm9.34$ & $3.51\pm4.10$ & $73.58\pm10.84$ & $3.50\pm5.11$ & $71.38\pm11.66$ & $4.42\pm5.71$ \\
\addlinespace[2pt]

U-Mamba~\cite{ma2024u} &
$81.71\pm8.70$ & $2.52\pm3.48$ & $80.90\pm8.26$ & $2.98\pm3.77$ & $79.49\pm9.17$ & $3.82\pm4.83$ & $78.06\pm10.47$ & $4.11\pm4.53$ \\
\addlinespace[2pt]

\midrule
\multicolumn{9}{c}{\textbf{State-of-The-Art Multimodal Methods}} \\
\midrule
\addlinespace[1pt]

AVSegFormer*~\cite{gao2024avsegformer} &
$83.68\pm7.94$ & $2.80\pm4.68$ & $81.27\pm6.40$ & $\underline{2.39\pm4.32}$ & $80.53\pm8.73$ & $\underline{3.48\pm4.77}$ & $76.59\pm9.16$ & $\mathbf{3.02\pm4.93}$ \\
\addlinespace[2pt]

AV-SAM*~\cite{mo2023av} &
$82.27\pm7.14$ & $2.38\pm2.83$ & $83.47\pm7.64$ & $3.01\pm2.19$ & $\underline{81.21\pm8.52}$ & $3.69\pm3.49$ & $\underline{80.62\pm8.23}$ & $4.26\pm4.88$ \\
\addlinespace[2pt]

Jain et al.~\cite{jain2024multimodal} &
$\underline{85.35\pm7.25} $ & $\underline{2.21\pm3.76}$ & $\underline{84.68\pm8.33}$ & $2.50\pm3.58$ & $81.19\pm9.36$ & $5.54\pm3.12$ & $80.14\pm10.42$ & $3.50\pm3.41$ \\
\addlinespace[2pt]

\midrule
\textbf{Ours} &
$\mathbf{90.82\pm6.30}$ & $\mathbf{1.48\pm1.72}$ & $\mathbf{87.74\pm6.41}$ & $\mathbf{2.11\pm2.43}$ & $\mathbf{86.37\pm7.04}$ & $\mathbf{2.36\pm2.90}$ & $\mathbf{83.78\pm8.25}$ & \underline{$3.43\pm3.17$} \\
\bottomrule
\multicolumn{9}{l}{\footnotesize{*Adapted.}} 
\end{tabular}%
}
\end{table*}

\subsection{Implementation Details}
All experiments are implemented in PyTorch 2.5.1~\cite{paszke2019pytorch} and conducted on an NVIDIA RTX A100 GPU with a batch size of 8. In Stage~2, the model is trained for 50 epochs using AdamW with weight decay $1\times10^{-2}$. During phase 1 (epochs 1-20), both encoders are frozen and only the fusion block and projection heads are optimized at a learning rate of $1\times10^{-4}$. During phase 2 (epochs 21--50), ViT layers are progressively unfrozen from top to bottom: Layer~12 at epoch 21, Layers~9--12 at epoch 31, each with a reduced learning rate of $1\times10^{-5}$ to preserve pretrained representations. In Stage~3, the pretrained ViT encoder, WavLM encoder, and fusion block are frozen to preserve the cross-modal representations learned in Stage~2. Only the newly introduced cross-attention layers and segmentation decoder are optimized using AdamW with a learning rate of $1\times10^{-4}$ and weight decay $1\times10^{-2}$. The segmentation loss combines cross-entropy and Dice losses with equal weights. Training proceeds for up to 100 epochs with early stopping (patience $=15$) based on validation Dice score.

\subsection{Performance Comparison}
We compare our method with state-of-the-art segmentation models, including both unimodal methods that rely solely on visual input and multimodal approaches. Evaluation metrics include Dice Coefficient (DSC) and Average Surface Distance (ASD). As shown in Tab.~\ref{tab:performance_comparison}, our method achieves leading performance on both datasets. On the most challenging (US-UT) setting, ours achieves further gains of $+6.88$\,DSC, and $-1.46$\,mm ASD; similar improvements are observed on USC-TIMIT ($+5.72$\,DSC). Notably, ours also outperforms recent foundation models including DINOv2~\cite{oquab2023dinov2} and MedSAM-2~\cite{zhu2024medical} by large margins. Among multimodal methods, ours consistently outperforms all competitors, achieving $+5.18$\,DSC and $-3.18$\,mm ASD over the strongest competitor Jain et al.~\cite{jain2024multimodal} on the US-UT configuration. On USC-TIMIT, although AVSegFormer achieves a marginally lower ASD, our method exhibits substantially smaller standard deviation, indicating more consistent and stable segmentation across subjects.


\begin{figure*}[t]
    \centering
    \includegraphics[width=0.85\textwidth]{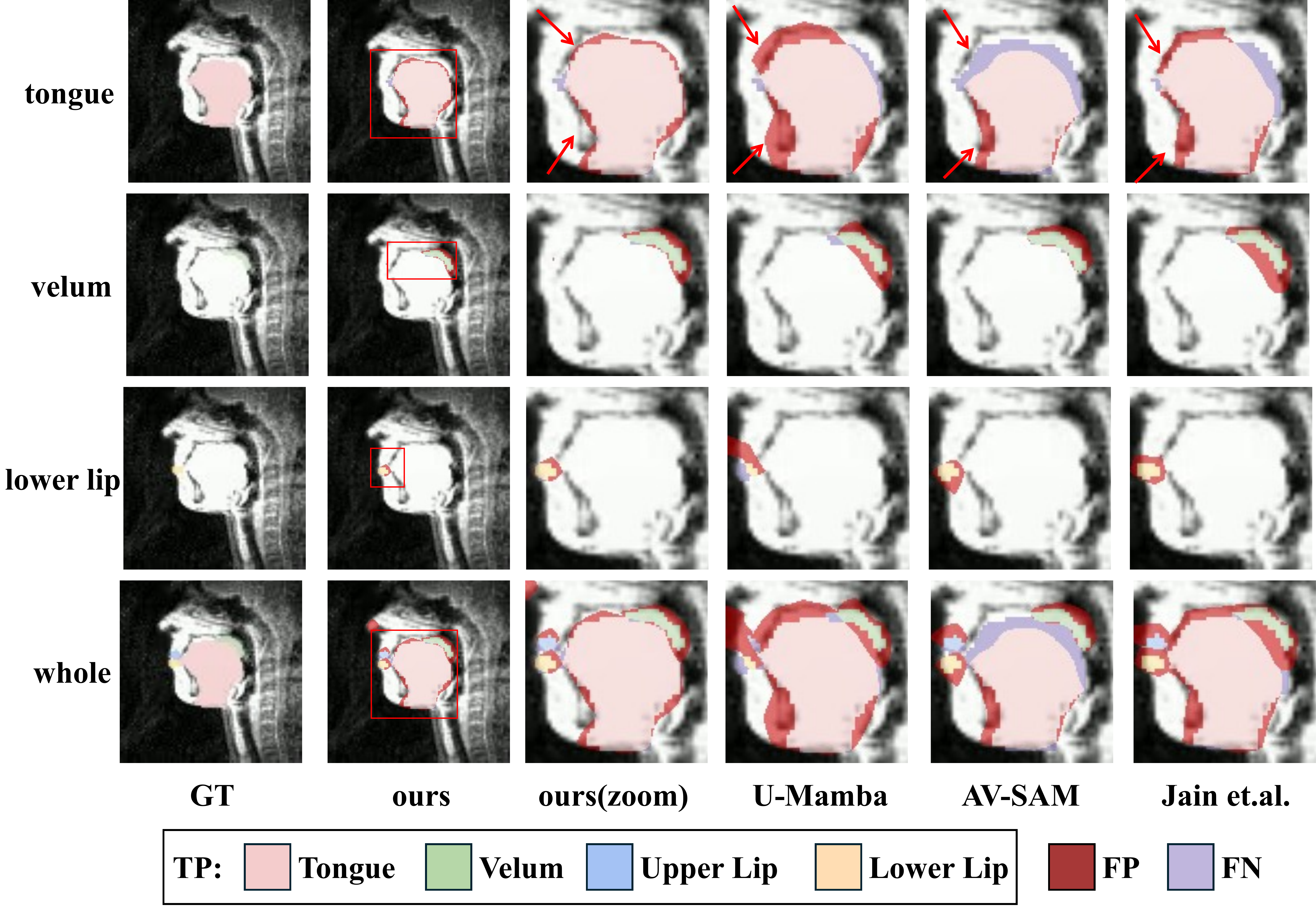}
    \caption{
    Qualitative comparison of articulator segmentation on a representative rtMRI frame. Colored regions in the figure indicate true positives for each articulator class. Red regions denote False Positive (FP), and purple regions denote False Negatives (FN). Red arrows highlight failure cases in competing methods. TP: True Positive.
    }
    \label{fig:qualitative}
\end{figure*}

Visually, as Fig.~\ref{fig:qualitative} shows, for the tongue (row 1), competing methods exhibit clear over-segmentation artifacts at both the superior and inferior boundaries (indicated by red arrows), incorrectly extending the predicted mask beyond the true tongue contour. In contrast, Our method closely adheres to the ground truth boundary. Ours also shows clear advantages in segmenting challenging small structures, such as the lower lip (row 3), yielding the lowest number of both false-positive and false-negative predictions.


\subsection{Ablation Study}
As shown in Tab.~\ref{tab:ablation} (A), we conduct ablation studies to verify the effectiveness of each proposed stage. We adopt the architecture of Jain et al.~\cite{jain2024multimodal} as baseline, which employs feature concatenation for cross-modal fusion. Introducing the audio modality IA Concat Fusion already yields a notable improvement of $+3.92$\,DSC and $-0.87$\,mm ASD over Image-only, confirming the complementary role of acoustic information in articulator segmentation. However, naively appending phonological features via concatenation IAP Concat Fusion provides only marginal gains ($+0.35$\,DSC). In contrast, converting phonological descriptors into spatial bounding-box priors +BBox Prior yields a substantial improvement of $+2.23$\,DSC and $-2.15$\,mm ASD, demonstrating the effectiveness of Stage~1 in bridging the modality gap. Progressively incorporating Stage~2 contrastive pretraining +Pretrain and Stage~3 cross-attention fusion brings further consistent gains across all metrics. Critically, comparing w/ full input ($1356$\,ms) and ours ($1058$\,ms)  reveals that image-only inference reduces latency by $22\%$ while incurring only a marginal performance gap ($-0.22$\,DSC, $+0.22$\,mm ASD), demonstrating that the cross-modal knowledge acquired during training is effectively distilled into the model weights. Tab.~\ref{tab:ablation} (B) presents the results of replacing the image encoder with alternative architectures while keeping all other components fixed. Although ResNet-34~\cite{he2016deep} achieves the fastest inference time ($751$\,ms) and reduces the total parameter count by $65.39$M, it yields a notable performance drop of $-4.11$\,DSC and $+2.75$ mm ASD compared to ViT-B~\cite{dosovitskiy2020image}. Swin-B~\cite{liu2021swin}, DINOv2-B~\cite{oquab2023dinov2} achieve a similar parameter count, but the model with ViT-B encoder achieves the best trade-off between performance, parameter efficiency, and inference speed.

\begin{table}[t]
\centering
\caption{Ablation study of our method on the 75-Speaker Annot-16 dataset under the US-UT configuration. (A)~Effect of each stage: ``Image-only'' denotes the baseline using image input solely. Inference time is measured in ms per 50 frames.(B) Effect of different image encoders. The best results are highlighted in \textbf{bold}, and the second-best are \underline{underlined}.  I: Image. A: Audio. Phon and P: Phonological class.}
\label{tab:ablation}
\setlength{\tabcolsep}{4pt}
\fontsize{8}{9}\selectfont
\resizebox{\columnwidth}{!}{%
\begin{tabular}{cccc ccc cccc}
\toprule
\multicolumn{11}{c}{\textbf{(A) Ablation Study of Different Stages}} \\
\midrule
\multicolumn{4}{c}{\textbf{Method}} & \multicolumn{3}{c}{\textbf{Input (Inference)}} & \multirow{2}{*}{\textbf{DSC(\%)}$\uparrow$} & \multirow{2}{*}{\textbf{ASD(mm)}$\downarrow$} & \multirow{2}{*}{\textbf{\#Params(M)}} & \multirow{2}{*}{\textbf{Time(ms)}$\downarrow$}  \\
\cmidrule(lr){1-4}\cmidrule(lr){5-7}
Name & Stage~1 & Stage~2 & Stage~3 & Image & Audio & Phon & & & &\\
\midrule
Image-only & \texttimes & \texttimes & \texttimes & \checkmark & \texttimes & \texttimes & $77.29\pm14.59$ & $6.41\pm5.22$ & $\mathbf{109.77}$ & $\mathbf{510}$\\
\addlinespace[1pt]
IA Concat Fusion & \texttimes & \texttimes & \texttimes & \checkmark & \checkmark & \texttimes & $81.21\pm9.36$ & $5.54\pm3.12$ & $\underline{161.53}$& $1173$ \\
\addlinespace[1pt]
IAP Concat Fusion & \texttimes & \texttimes & \texttimes & \checkmark & \checkmark & \checkmark & $81.56\pm8.71$ & $6.03\pm3.61$ & $162.27$ & $1214$\\
\addlinespace[1pt]
+ BBox Prior & \checkmark & \texttimes & \texttimes & \checkmark & \texttimes & \texttimes & $83.79\pm10.31$ & $3.88\pm4.51$ & $\underline{161.53}$ & $1008$\\
\addlinespace[1pt]
+ Pretrain & \checkmark & \checkmark & \texttimes & \checkmark & \texttimes & \texttimes & $84.70\pm9.18$ & $3.64\pm3.77$ & $163.10$ &$\underline{ 965}$\\
\addlinespace[1pt]
~w/~full~input & \checkmark & \checkmark & \texttimes & \checkmark & \checkmark & \checkmark & $\mathbf{86.59\pm6.32}$ & $\mathbf{2.14\pm3.13}$ & $186.64$& $1356 $\\
\addlinespace[1pt]
ours & \checkmark & \checkmark & \checkmark & \checkmark & \texttimes & \texttimes & $\underline{86.37\pm7.04 }$& $\underline{2.36\pm2.90} $& $186.64$& $1058 $\\
\midrule
\multicolumn{10}{c}{\textbf{(B) Ablation Study of Image Encoder}} \\
\midrule
\multicolumn{7}{c}{\textbf{Image Encoder}} & \textbf{DSC(\%)}$\uparrow$ & \textbf{ASD(mm)}$\downarrow$ & \textbf{\#Params(M)} & {\textbf{Time(ms)}$\downarrow$} \\
\midrule
 \multicolumn{7}{c}{ResNet-34~\cite{he2016deep}} & $82.26\pm5.97$ & $5.11\pm3.28$ & $\mathbf{121.25}$ & $\mathbf{751}$\\
\addlinespace[1pt]
\multicolumn{7}{c}{Swin-B~\cite{liu2021swin}}       & $83.11\pm9.51$ & $4.36\pm4.14$ & $188.92$& $1245$ \\
\addlinespace[1pt]
\multicolumn{7}{c}{DINOv2-B~\cite{oquab2023dinov2}}          & $\underline{84.03\pm8.43}$ & $\underline{3.55\pm4.03}$ & $\underline{186.64}$ &$ 1072$\\
\addlinespace[1pt]
\multicolumn{7}{c}{ViT-B (Ours)~\cite{dosovitskiy2020image}}        & $\mathbf{86.37\pm7.04}$& $\mathbf{2.36\pm2.90}$ & $\underline{186.64}$ &$ \underline{1058}$\\
\bottomrule
\end{tabular}%
}
\end{table}

\section{Discussion \& Conclusion}
We present the first framework to combine visual and acoustic information for vocal tract articulator segmentation in rtMRI while requiring only image modality at inference time. By explicitly incorporating phonological class priors as spatial bounding-box maps, our method bridges the modality gap between symbolic linguistic knowledge and spatial image representations. By pretraining the encoders via contrastive learning, the learned cross-modal representations are fully encoded into the model weights, making the framework robust to audio absence in clinical deployment. Furthermore, audio-guided cross-attention allows visual tokens to dynamically attend to temporally resolved acoustic features, enabling fine-grained integration of articulatory motion patterns. These results demonstrate that structured phonological supervision and cross-modal alignment provide complementary and transferable benefits that persist at inference even without audio input.

Nevertheless, there is still room for improvement. The current framework processes each frame independently, without modeling temporal dependencies across frames, which may introduce jitter in predicted trajectories. Furthermore, phonological features are treated as context-free labels, neglecting coarticulation effects that systematically shape articulator positions based on neighboring phonemes. Finally, all experiments are conducted on healthy speaker data due to the scarcity of annotated pathological rtMRI datasets. Future work should explore temporal architectures, context-aware phoneme embeddings derived from speech language models, and validation on dysarthric and post-surgical speech data to fully assess the clinical utility of the proposed framework.


%
%
%

%




\end{document}